\documentclass[11pt]{article}

\usepackage[preprint]{acl}

\usepackage{times}
\usepackage{latexsym}

\usepackage[T2A,T1]{fontenc}
\usepackage[utf8]{inputenc}

\usepackage[ukrainian,english]{babel}

\usepackage{tempora}

\usepackage{microtype}
\usepackage{xcolor}
\usepackage{tcolorbox}
\usepackage{booktabs}
\usepackage{multirow}
\usepackage{array}
\usepackage{tabularx}
\usepackage{inconsolata}
\usepackage{placeins}
\usepackage{float}
\usepackage{fontawesome5}
\usepackage{pifont}
\newcommand{\cmark}{\textcolor{green!60!black}{\ding{51}}}  % checkmark
\newcommand{\xmark}{\textcolor{red}{\ding{55}}}             % x mark

\usepackage{graphicx}
\usepackage{xcolor}
\title{Recovered in Translation: Efficient Pipeline for Automated Translation of Benchmarks and Datasets}

\author{
  \textbf{Hanna Yukhymenko\textsuperscript{1$\dagger$, 2}},
  \textbf{Anton Alexandrov\textsuperscript{1}},
  \textbf{Martin Vechev\textsuperscript{1,2}}
\\
\\
  \textsuperscript{1}INSAIT, Sofia University "St. Kliment Ohridski",
  \textsuperscript{2}ETH Zurich
\\
  \small{
    \textbf{Correspondence:} \href{mailto:hyukhymenko@ethz.ch}{hanna.yukhymenko@insait.ai}
  }
\\
\\
  {
    \faGithub~\textbf{Code:} \href{https://github.com/insait-institute/ritranslation}{insait-institute/ritranslation}
  }
\\
  {
    \includegraphics[height=1em]{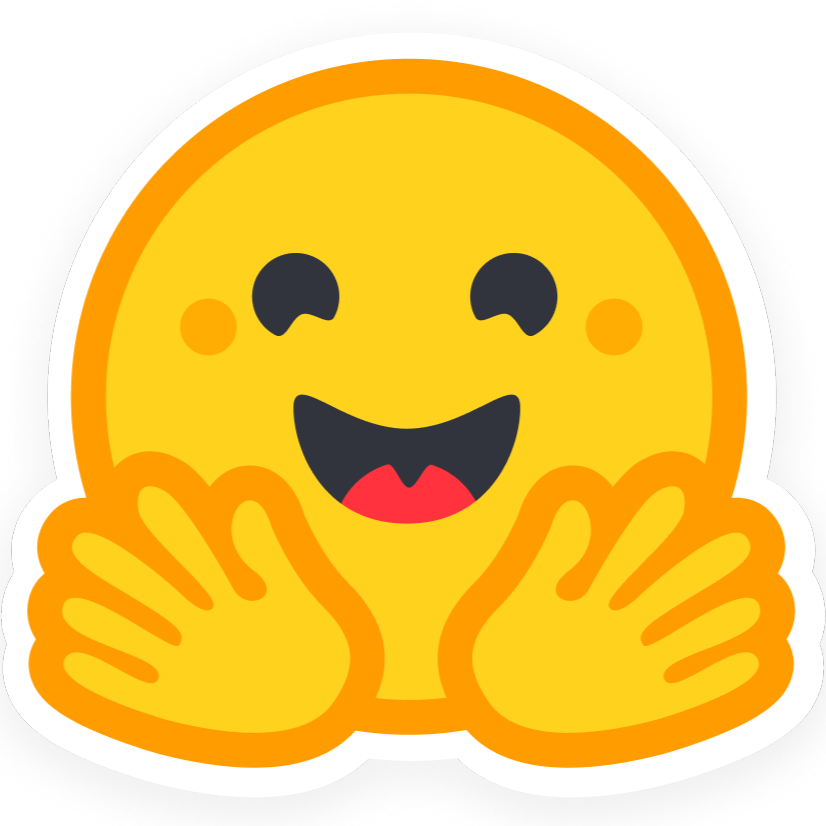}~\textbf{Benchmarks:} \href{https://huggingface.co/collections/INSAIT-Institute/multilingual-benchmarks}{insait-institute/multilingual-benchmarks}
  }
}

\begin{document}
\maketitle
\setcounter{footnote}{0}
\footnotetext{Under review. Preprint version.}
\renewcommand{\thefootnote}{\fnsymbol{footnote}}
\setcounter{footnote}{2}
\footnotetext{Work done while at INSAIT}
\begin{abstract}

The reliability of multilingual Large Language Model (LLM) evaluation is currently compromised by the inconsistent quality of translated benchmarks. Existing resources often suffer from semantic drift and context loss, which can lead to misleading performance metrics. In this work, we present a fully automated framework designed to address these challenges by enabling scalable, high-quality translation of datasets and benchmarks. We demonstrate that adapting test-time compute scaling strategies, specifically Universal Self-Improvement (USI) and our proposed multi-round ranking method, T-RANK, allows for significantly higher quality outputs compared to traditional pipelines. Our framework ensures that benchmarks preserve their original task structure and linguistic nuances during localization. We apply this approach to translate popular benchmarks and datasets into eight Eastern and Southern European languages (Ukrainian, Bulgarian, Slovak, Romanian, Lithuanian, Estonian, Turkish, Greek). Evaluations using both reference-based metrics and LLM-as-a-judge show that our translations surpass existing resources, resulting in more accurate downstream model assessment. We release both the framework and the improved benchmarks to facilitate robust and reproducible multilingual AI development.
\end{abstract}

\section{Introduction}

Large Language Models have demonstrated rapid progress in machine translation, outperforming classical tools such as Google Translate and DeepL \citep{deutsch2025wmt24pp}. This advancement benefits multilingual model development by addressing data and benchmark scarcity in non-English settings. Recent work shows that sampling multiple translation candidates at higher temperatures enhances translation quality, with test-time scaling methods like Best-of-N \citep{stiennon2020learning} and Fusion-of-N \citep{khairi2025makingtakingbestn} demonstrating improved performance across multilingual domains.

\paragraph{Key challenge: multilingual benchmark quality.} A significant gap in recent research is the limited exploration of benchmark translation quality. Multilingual benchmarks are critical for assessing the quality of given model, yet receive less attention than data collection. Existing benchmarks remain imperfect due to reliance on older LLMs or oversimplified methods -- most were translated using classical MT tools lacking instruction prompting capabilities, or via older models like GPT-4 whose multilingual capabilities lag behind current frontier models \citep{lai2023okapi}.

We investigate available benchmarks and identify common translation flaws stemming from translating questions and answers separately, leading to contextual and grammatical mismatches. We focus on Eastern and Southern European languages because: (1) they exhibit complex grammatical features (extensive case systems, grammatical gender, aspect-based verbs) that are sensitive to contextual misalignment; (2) they represent mid-resource languages where frontier models still lack adequate localization; (3) unlike lower-resource languages, translated benchmarks exist for comparison. We translate MMLU, Hellaswag, ARC, and Winogrande into Ukrainian, Romanian, Slovak, Lithuanian, Bulgarian, Turkish, Greek and Estonian.

In this paper, we address three key research questions:

\begin{itemize}
  \item \textbf{RQ1}: Are current translated multilingual benchmarks robust and reliable?
  \item \textbf{RQ2}: To what extent do test-time compute scaling methods improve machine translation quality?
  \item \textbf{RQ3}: How can we translate benchmarks and datasets into other languages while efficiently integrating their unique linguistic features into the evaluated tasks?
\end{itemize}

\paragraph{Our work:} To address these questions, we present a novel automated translation framework supporting four methods across various model types, including open-weight models. The framework facilitates machine translation of datasets and benchmarks with minimal manual supervision and maximum configurability, offering multiple methods to balance costs and time investment based on language performance in selected MT models. Our framework incorporates: (1) classic one-prompt translation with optional self-correction (SC) as a lightweight solution; (2) Best-of-N with LLM-as-a-judge scoring to provide reward signals; (3) Universal Self-Improvement (USI) with multi-prompt sampling and unification of translation candidates into refined outputs; (4) our newly proposed Translation Ranking (T-RANK), which employs multi-prompt candidate sampling and multi-round competitive ranking to enhance error detection and achieve superior translation quality. 

We evaluate these test-time scaling methods across several mid-resource languages, translate popular benchmarks into these languages, and compare our results with existing translations. Our findings demonstrate that the integrated methods produce higher-quality translations with more accurate evaluation results and improved text quality. Moreover, these methods yield performance improvements on established machine translation benchmarks such as WMT24++ and FLORES.

\paragraph{Main contributions:} We summarize our key contributions as follows:

\begin{itemize}
  \item A comprehensive analysis of multilingual benchmark quality, identifying the root causes of current translation deficiencies.
  \item A novel, fully automated framework that enables efficient translation of datasets and benchmarks with maximum flexibility. The pipeline is fully configurable, allowing practitioners to optimize the balance between translation quality and cost in multilingual development.
  \item We release several benchmarks translated into Eastern and Southern European languages, including Ukrainian, Romanian, Slovak, Lithuanian, Bulgarian, Turkish, Greek and Estonian. We find, that translation quality influences benchmark evaluation results significantly, and our proposed methods yield more reliable and accurate multilingual evaluation.
\end{itemize}

These findings enable more efficient multilingual model development by improving both data and benchmark acquisition.

\section{Background and Related Work}

\subsection{Foundation for LLM-Based Efficient Translation}

The rapid expansion of multilingual applications for large language models (LLMs) necessitates scalable and high-quality translation frameworks. Recent advancements propose innovative methodologies to enhance LLM-based translation through adaptive prompting and self-refinement strategies.

The Adaptive Few-shot Prompting (AFSP) framework addresses prompt sensitivity in machine translation by dynamically selecting suitable translation demonstrations. AFSP retrieves semantically similar examples from parallel corpora based on LLM-generated embeddings, then generates and reranks multiple translation candidates to select the most contextually appropriate translation \citep{tang2025adaptive}.

The TEaR (Translate, Estimate, and Refine) framework introduces systematic self-refinement for LLM-based translation. TEaR operates by translating the source text, estimating translation quality, and refining based on identified errors. This iterative process enables LLMs to self-correct and improve translation quality, with cross-model experiments demonstrating that general-purpose LLMs can perform both translation and evaluation simultaneously \citep{feng2024tear}.

In addition to these frameworks, several test-time compute scaling methods show potential for enhancing LLM translation capabilities. Although originally designed for general reasoning tasks such as mathematics and coding, these methods demonstrate promising applications in translation:

\begin{itemize}
  \item \textbf{Best-of-N Sampling}: This method generates multiple translation outputs and selects the best one based on predefined criteria. It leverages the diversity in LLM outputs due to temperature variations, increasing the likelihood of obtaining high-quality translations \citep{stiennon2020learning}.
  
  \item \textbf{Universal Self-Consistency (USC)}: USC extends the concept of self-consistency by enabling LLMs to select the most consistent answer among multiple candidates without relying on answer extraction processes. This approach is particularly effective for open-ended generation tasks, improving performance in mathematical reasoning, code generation, and summarization \citep{chen2023universal}.

  \item \textbf{Fusion-of-N}: Rather than selecting the single best answer as in Best-of-N, Fusion-of-N synthesizes the most informative elements from multiple candidates into a single final output. The method uses an LLM judge to aggregate diverse strengths from the candidate pool. Fusion-of-N outperforms Best-of-N and shows promising gains on multilingual tasks, including machine translation \citep{khairi2025makingtakingbestn}.
\end{itemize}

Findings from \citet{khairi2025lifegivessamplesbenefits} also confirm, that sampling multiple candidates with higher temperature with self-improvement and refined selection process leads to significant improvements in many multilingual domains, including machine translation.

Results from the WMT24++ benchmark \citep{deutsch2025wmt24pp} further corroborate the efficacy of these methodologies, demonstrating that state-of-the-art LLMs outperform traditional machine translation tools across various language pairs. The research suggests that integrating test-time compute strategies can significantly enhance translation quality, particularly for mid- or low-resource languages. A clear trend emerges whereby newer and larger LLMs consistently outperform existing tools and earlier model iterations, indicating the potential for continued improvements in translation quality as more advanced models are released.

\begin{takeaway}
\textbf{Takeaway 1:} Large language models outperform traditional machine translation systems like Google Translate and DeepL through targeted prompts that address domain-specific and language-specific translation challenges.
\end{takeaway}

Collectively, these methodologies underscore the importance of adaptive prompting, self-refinement, and test-time scaling strategies in enhancing LLM-based translation systems. By integrating these approaches, it is possible to develop more robust, efficient, and high-quality translation frameworks capable of addressing the challenges posed by mid- and low-resource languages while minimizing the need for manual human intervention.

\subsection{Shortcomings of Existing Multilingual Resources}

The rapid development of large language models (LLMs) has highlighted the need for robust multilingual benchmarks. However, translating existing benchmarks often introduces issues that compromise accurate assessment of LLM capabilities across languages.

A prominent example is the MuBench benchmark dataset introduced by \citet{han2025mubenchassessmentmultilingualcapabilities}, comprising widely used benchmarks (Hellaswag \citep{zellers2019hellaswagmachinereallyfinish}, ARC \citep{clark2018thinksolvedquestionanswering}, Winogrande \citep{sakaguchi2019winograndeadversarialwinogradschema}, MMLU \citep{hendrycks2021measuringmassivemultitasklanguage}, and others) translated into 61 languages with 3.9M samples. While this initiative represents an important step in multilingual evaluation, the translation process relies on an automated pipeline with quality control measures including semantic consistency evaluation, translation purity assessment, and cultural sensitivity checks. However, despite human expert evaluation of 30k samples across 15 languages combined with LLM-as-a-Judge validation, the predominantly automated approach raises concerns regarding translation quality for the remaining samples, as machine translation methods may fail to capture nuanced linguistic and cultural contexts in all 61 languages. Further variations in quality assurance across languages could introduce semantic inaccuracies that skew evaluation results.

Beyond general quality concerns, certain benchmarks present inherent translation challenges. For example, Winogrande, MMLU and Hellaswag may contain answer options with gender-specific adjective endings for sentence completion tasks. In languages with gender-specific adjectives, translating these options can inadvertently reveal correct answers or mislead towards picking an incorrect option, allowing models to succeed through language proficiency rather than reasoning capability. Such linguistic features can compromise evaluation integrity through unintended answer leakage. Therefore, there is a critical need for a flexible framework which allows to adapt translation methods and prompts to address specific issues for different benchmark and language combinations, while MuBench universal approach lacks such flexibility and does not explicitly address these challenges.

The Global-MMLU project \citep{singh2024globalmmlu} represents a substantial effort to advance multilingual evaluation by translating the MMLU benchmark into 42 languages. The process combines machine translation using Google Translate with crowdsourced human verification; however, only approximately 20\% of machine-translated texts underwent manual correction. Moreover, questions and answers were translated separately, resulting in observable grammatical inconsistencies in translated entries for languages such as Ukrainian. This context-agnostic translation approach can produce inconsistencies or grammatical errors that affect benchmark coherence and accuracy, particularly for lower-resource languages. Additionally, relying on tools like Google Translate without thorough human validation may overlook language-specific nuances essential for accurate evaluation.

The Okapi framework \citep{lai2023okapi} introduces a novel approach to multilingual instruction tuning by leveraging Reinforcement Learning from Human Feedback (RLHF). Unlike traditional methods relying solely on supervised fine-tuning (SFT), Okapi combines SFT with RLHF to align model outputs more closely with human preferences across diverse languages. Moreover, some of the popular benchmarks like MMLU, Hellaswag and ARC were also translated to 26 languages to enable multilingual evaluation. However, Okapi's translation process utilized the ChatGPT model series and the translation could be further improved by employing test-time compute scaling methods to enhance translation quality. Additionally, the framework does not explicitly address language-specific grammatical features that may impact benchmark coherence and accuracy. We present the examples of such translation flaws mentioned before in Appendix~\ref{app:translation_examples}.

\begin{takeaway}
\textbf{Takeaway 2:} Translating questions and answer options within the same prompt context is essential for sentence completion tasks, as it preserves semantic relationships and prevents contextual mistakes during evaluation.
\end{takeaway}

Recent studies, including WMT24++ \citep{deutsch2025wmt24pp}, demonstrate that state-of-the-art LLMs outperform traditional machine translation tools such as DeepL or Google Translate across all evaluated language pairs. This finding suggests that reliance on older translation tools may be insufficient for ensuring high-quality translations in multilingual datasets and benchmarks. In addition, practical limitations of popular frameworks (such as low API rate limits or inability to use advanced prompting techniques) can hinder large-scale translation efforts, making it more difficult to produce comprehensive and accurate datasets.

In summary, while initiatives such as MuBench, Global-MMLU, and Okapi represent important progress toward multilingual LLM evaluation, the methods employed in translating benchmarks exhibit limitations that hinder improvement. Dependence on automatic translation without comprehensive human verification, challenges arising from language-specific grammatical structures, and the limited capabilities of translation tools collectively highlight the need for more sophisticated approaches that respect linguistic features. Further, current frontier language models are already outperfoming classical tools, which were used to translate some of the widely used multilingual benchmarks. Future research should advance automated methodologies that reduce reliance on manual curation while simultaneously improving both the quality and accessibility of multilingual evaluation resources.

\section{An Efficient Automated Translation Framework}

In this section we present a novel automated translation framework, which utilizes Large Language Models with features adapted for both dataset and benchmark (QA/test) formats. Moreover, the framework also allows flexibility in methods and their inner configurable parameters to optimize cost- and time-effectiveness.

\subsection{Motivation}

As highlighted in previous sections, we observed the lack of research and solutions towards scaled automated translation adapted for custom benchmark formats without loss of translation quality. For this framework, we aim to tackle the following key problems in LLM-assisted data translation:

\begin{itemize}
  \item \textbf{Support for Diverse Data Formats}: Many datasets have complex, nested structures that complicate processing. For LLM training, flat string fields without hierarchical nesting ensure easier ingestion and more efficient workflows.
  \item \textbf{Preservation of Benchmark QA Structure}: Benchmark evaluation reliability depends on maintaining coherence between questions and answer choices. Preserving these relationships during translation ensures benchmarks remain faithful to the original task structure.
  \item \textbf{Adapting to Varying Language Resource Availability}: High-resource languages (German, Spanish, Hindi) achieve strong translation quality with simple zero-shot prompting, while mid- and low-resource languages require sophisticated, language-specific pipelines. Users need flexibility to configure translation approaches based on each language's characteristics.
  \item \textbf{Addressing Language-Specific Phenomena}: EEU languages possess unique grammatical features requiring careful handling. Language-specific prompt engineering with few-shot examples and LLM-powered verification stages can improve translation fidelity by identifying and correcting language-specific errors.
\end{itemize}

\subsection{Methodology}

In our framework, we propose two configuration modes -- Dataset and Benchmark. They use different prompts and data format handling, since dataset is more straightforward, but benchmark has a complex connection between question and answers, which needs to be preserved. Finally, we propose four translation methods:

\begin{itemize}
  \item \textbf{SC (Self-Check)}: 0-shot simple translation with optional additional check from another judge LLM.
  \item \textbf{Best-of-N sampling}: sampling N translation candidates and prompting the model to score them, then picking the one with the highest score.
  \item \textbf{USI (Universal Self-Improvement)}: sampling N translations, asking the model to combine them into the best one according to predefined evaluation criteria.
  \item \textbf{T-RANK (Translation Ranking)}: sampling N translations, asking the model to rank them according to their quality and coherence, then correcting the best candidate.
\end{itemize}

We now examine the translation methods employed in our framework, highlighting their advantages and optimal use cases.

\textbf{Default translation with optional self-check (SC).} This method involves a simple 0-shot prompting of the LLM to translate a text. The user has an option to include a self-check stage: after translation, the model (in a new chat with no history) is prompted to evaluate and correct the result with respect to the original text content. This method is suitable for large text translation using high-resource languages, as it is less costly to perform, due to the sufficient translation capabilities in high-resource languages; however, since the model is prompted to look out for potential errors it might hallucinate them. Additionally, there exists an option to include a few-shot prompt to provide an example of which language-specific points the model should consider during translation.

\textbf{Best-of-N sampling (BoN).} Drawing from test-time compute scaling methods, we implement Best-of-N sampling without a reward model to maintain a training-free framework. We sample N translations at higher temperature (0.7) for diversity, then prompt the LLM to score candidates 1-10 based on specified criteria (Appendix~\ref{app:translation_prompts}), selecting the highest-scoring translation. While cost-effective and language-agnostic, this method yields lower quality than T-RANK and USI, as LLMs exhibit limitations in numerical scoring and positional bias, favoring earlier candidates despite identifying obvious errors.

\begin{figure}[h]
  \includegraphics[width=\columnwidth]{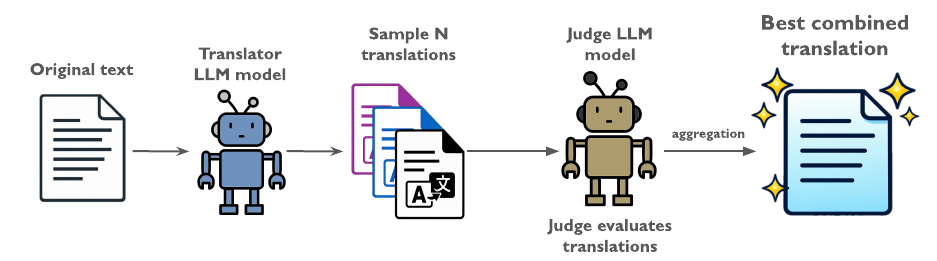}
  \caption{Universal Self-Improvement (USI) method workflow outlook}
  \label{fig:usi_method_fig}
\end{figure}

\textbf{Universal Self-Improvement (USI).} Building on Universal Self-Consistency and Fusion-of-N, this method operates on the principle that the most consistent translation is not necessarily the best. While Fusion-of-N improves translation quality substantially, the absence of precise translation-specific metrics limits the model's ability to efficiently identify each candidate's strengths for optimal merging, potentially resulting in fusion outputs that incorporate errors. Therefore, we adapt this approach to cultivate self-improvement specifically for machine translation as shown in Appendix~\ref{app:translation_examples} Figure~\ref{fig:usi_correction}. We sample N candidate translations using higher temperature (0.7 recommended), then present them to an evaluator LLM with instructions to combine the candidates into the best version according to specified criteria (shown in Appendix~\ref{app:translation_prompts} Table~\ref{tab:translation_prompts}). During a combination phase, we ask the model to use only the best features from all seen candidates. This method proves cost-efficient and time-efficient, requiring only $N+1$ model calls per entry, while successfully addressing and correcting language-specific features. We illustrate the Universal Self-Improvement workflow in more detail in Figure~\ref{fig:usi_method_fig}.

\textbf{Translation Ranking (T-RANK).} Building upon prior work in translation ranking, we propose an adaptive approach that refines this concept. This method begins by sampling N diverse translations with high temperature. A judge model then evaluates these candidates by ranking them according to predetermined quality criteria, checking general quality, domain consistency, preservation of original question idea and other (shown in Appendix~\ref{app:translation_prompts} Table~\ref{tab:translation_prompts}), with the top-ranked translation selected as the final output, optionally with refinements. Using this method, we observe more sophisticated reasoning from non-reasoning models, as they successfully identify subtle errors that other methods fail to correct. Moreover, evaluation through comparative ranking rather than numerical scoring appears to facilitate more detailed and rigorous assessment of translation quality.

\begin{figure}[h]
  \includegraphics[width=\columnwidth]{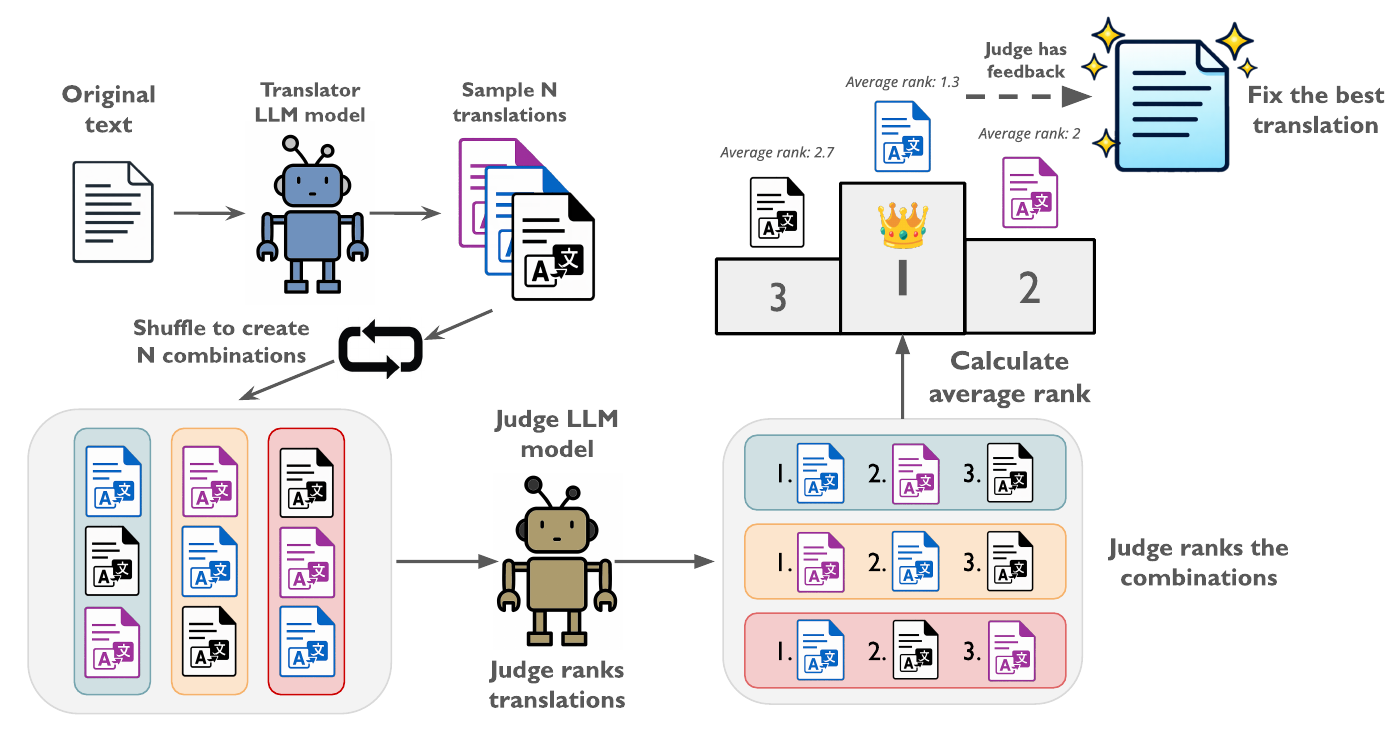}
  \caption{Translation Ranking (T-RANK) method workflow outlook}
  \label{fig:trank_method_fig}
\end{figure}

Previous research on LLM-based evaluation has identified several risks and biases inherent to such methodologies. One notable issue is positional bias: LLMs tend to assign higher scores to candidates presented earlier in the sequence (as shown in Appendix~\ref{app:translation_examples} Table~\ref{tab:trank_positional_bias}). Also, LLM judges often focus their evaluation on the first candidate, identifying specific flaws, and subsequently bias their assessment of following candidates by primarily searching for similar errors, potentially overlooking novel issues in later candidates.

To reduce these prejudices, we introduce a multiple-round sequential ranking strategy, in which the candidate translations are systematically moved across different positional orders. In particular, each translation candidate is presented once in every possible position across all rounds. This process can be visualized as the construction of a sequential grid, where with each new round, candidates are moved one position further till the end (as visualized in Figure~\ref{fig:trank_method_fig}). After $N$ rounds for $N$ candidates, each candidate has been presented in each position exactly once. Lastly, we show all the candidates to the judge model again and ask to correct and refine the selected translation candidate if needed -- this improves model's ability to notice potential translation flaws by looking at strong and weak sides of other variants. We noticed an improved behavior and reasoning when using this method and in particular, presenting \emph{all} candidates to the judge model each time when doing the final correction, as shown in Appendix~\ref{app:translation_examples} Figure~\ref{fig:trank_correction}. This approach demonstrates a promising improvement in reliability of the evaluation by reducing positional biases while controlling evaluation costs and computational efficiency with a total of $2N+1$ model calls. Optionally, once can combine best candidate correction during ranking, however, we found evaluator model's reasoning to be more attentive in a separated setup. Compared to output of USI method shown in Appendix~\ref{app:translation_examples} Figure~\ref{fig:trank_correction}, we can observe that USI method performs fusion of methods, while T-RANK finds more inconsistencies across all candidates and tries to remove them. Since USI fusion step is done only once, the judge model tends to focus on translation flaws seen in first candidates and then tries to check for those specific errors in other candidates. With the multi-round ranking the model is able to see all possible pitfalls from presented options, improving the quality of selection and final improvement even further, while mitigating positional bias at the same time.

We provide example prompts for all proposed methods in Appendix~\ref{app:translation_prompts}.

\begin{takeaway}
\textbf{Takeaway 3:} While Best-of-N and USI sampling improve translation quality, the T-RANK method's competitive ranking approach proves more efficient at identifying subtle translation errors.
\end{takeaway}

\subsection{Benchmark Translation for Eastern European Languages}

We have translated several popular benchmarks into Eastern European languages using our proposed framework. The selected benchmarks include MMLU, Hellaswag, ARC, and Winogrande, which are widely used for evaluating LLM capabilities across various tasks. We focused on Eastern and Southern European languages such as Ukrainian, Romanian, Slovak, Lithuanian, Greek, Bulgarian, Turkish and Estonian due to their complex grammatical structures and mid-resource status. These languages present unique challenges for machine translation, making them ideal candidates for testing the effectiveness of our framework. MMLU was translated using the GPT-4o-mini-2024-07-18 model checkpoint from OpenAI and all mentioned benchmarks were translated to Ukrainian using the same model. Remaining benchmark translations to other languages were performed using the Gemini-2.0-Flash model from Google. We provide a detailed breakdown of used translation models and methods in Appendix~\ref{app:bench_stats} Table~\ref{tab:best_models}. In the next section and in Appendix~\ref{app:evals}, we present a comprehensive evaluation of the translation quality achieved through our framework, comparing it with existing translations and assessing its impact on benchmark performance.

\section{Evaluation Results}

\subsection{Machine Translation Benchmarks}

We utilize two datasets to evaluate the translation quality of our proposed methods: FLORES and WMT24++.

\textbf{FLORES}. The FLORES benchmark evaluates machine translation systems across multilingual scenarios. FLORES-101 provides 3,001 (1,012 devtest split) sentences from English Wikipedia professionally translated into 101 languages, enabling evaluation of many-to-many translation systems, particularly for low-resource language pairs \citep{guzman2019flores, goyal2022flores101}.

\textbf{WMT24++}. The WMT24++ benchmark covers 55 languages and dialects with human-written references and post-edits across four domains: literary, news, social, and speech. This diversity enables comprehensive evaluation of translation systems across high-resource and low-resource languages \citep{deutsch2025wmt24pp}.

Both FLORES and WMT24++ serve as standard benchmarks in the machine translation community due to their broad language coverage and high-quality human translations, enabling meaningful comparisons across models and approaches.

We evaluate our proposed methods on English-Ukrainian translation using the COMET (Crosslingual Optimized Metric for Evaluation of Translation) metric. COMET leverages multilingual pre-trained models to assess translations by comparing source text, hypothesis, and reference, demonstrating higher correlation with human judgments than traditional metrics like BLEU or chrF++ \citep{rei-etal-2020-comet}. We report COMET system-level scores (aggregated across all translations) in Table~\ref{tab:transsoup_methods_eval}, using the Unbabel/XCOMET-XL model for reference-based quality estimation \citep{guerreiro2023xcomettransparentmachinetranslation}. For mentioned tasks we have used GPT-4o-mini-2024-07-18 model for translation.

\begin{table}[h]
\centering
\setlength{\tabcolsep}{8pt}
\begin{tabular}{l cc}
\toprule
\textbf{Method} & \textbf{WMT24++} & \textbf{FLORES} \\
\midrule
Baseline & 0.827 & 0.937 \\
SC (with check) & 0.821 & 0.937 \\
Best-of-N (n=5) & 0.843 & 0.943 \\
USI (n=5) & 0.843 & \textbf{0.945} \\
T-RANK (p=5) & \textbf{0.845} & 0.940 \\
\bottomrule
\end{tabular}
\caption{COMET (reference-based) translation quality scores for WMT24++ and FLORES EN$\to$UK pair (GPT-4o-mini). $n$ denotes number of sampled candidates, $p$ denotes number of different translation prompts (for $p>1$ we use $n=1$).}
\label{tab:transsoup_methods_eval}
\end{table}

Our results indicate that USI and T-RANK demonstrate clear advantages over other methods; however, it remains unclear whether T-RANK consistently outperforms USI. This raises the question of whether a correlation exists between translation cost or effort and quality. We must account for the fact that COMET evaluations are not entirely reliable, even when reference translations are provided. Notably, both datasets used for evaluation (WMT24++ and FLORES) primarily contain short texts that do not adequately test complex grammatical structures, particularly for Ukrainian. Moreover, the choice of a correct translation often depends on personal preferences and stylistic conventions; a single source text may have multiple valid translation candidates, all grammatically correct. Therefore, reference translations in commonly used machine translation benchmarks cannot be considered absolute gold standards, and achieving the highest COMET scores does not guarantee that a method is error-free. This motivates the need for additional comparative tests under different evaluation paradigms, such as Quality Estimation (QE) or reference-free machine translation evaluation. As discussed in appendix~\ref{app:mt_benchmarks}, USI is more suitable for short and simple dataset translation, whereas T-RANK shows better performance when translating benchmarks, especially when they have complex question structure, which might get "lost in translation" for languages explored in this paper.

For reference-free QE, we can directly use COMET to score translations without requiring an ideal reference translation. We compared USI and T-RANK methods using the QE setup on FLORES and MMLU datasets, which contain longer and more complex texts. In this work, we use Unbabel/wmt23-cometkiwi-da-xl model for reference-free QE COMET evaluation \citet{rei2023scalingcometkiwiunbabelist2023} and the detailed evaluations are provided in appendix~\ref{app:mt_benchmarks}.

We should also note, that using COMET models for machine translation quality evaluation has certain limitations, including bias towards specific domains in training data, technical setup impacting reproducibility, and potential discrepancies with human judgment \citep{zouhar-etal-2024-pitfalls}. While COMET provides a useful automated metric, it is not infallible and should be complemented with human evaluation or using LLM-as-a-judge for comprehensive assessment.

\subsection{Multilingual Benchmark Translation Quality}

Machine translation quality estimation models struggle to provide consistent and justified feedback on bigger sized texts. Moreover, they are not to able to evaluate the preservation of questions and answer content for the benchmark format, Therefore, we also use LLM-as-a-judge to compare MMLU translations between standard industry-recognized source (Global-MMLU) and our proposed translation methods (T-RANK/USI). MMLU was translated using GPT family of models, thus, we select a model from another family, Gemini-2.5-Flash, as a judge, using which we showcase a clear advantage in translation quality using our proposed methods across Ukrainian, Romanian, and Lithuanian, as presented in Table~\ref{tab:trank_mmlu_judge}. Notably both methods outperform Global-MMLU alternative in quality of translations, confirming the effectiveness of our framework for benchmark translation. 

\begin{table}[h]
\centering
\setlength{\tabcolsep}{4pt}
\begin{tabular}{l l ccc}
\toprule
\textbf{Language} & \textbf{Method} & \textbf{Wins} & \textbf{Draws} & \textbf{Losses} \\
\midrule
UK & T-RANK & \textbf{8750} & 3276 & 2016 \\
RO & T-RANK & \textbf{8376} & 3020 & 2646 \\
LT & USI & \textbf{7382} & 3181 & 3478 \\
\bottomrule
\end{tabular}
\caption{LLM-as-a-judge comparison between Global-MMLU and our translations (T-RANK/USI) using Gemini-2.5-Flash as judge. Our translations win significantly more comparisons across all evaluated languages.}
\label{tab:trank_mmlu_judge}
\end{table}

We also compare evaluation results of mid-sized models like Gemma 3 4/12B, Qwen 3 8B and Llama 3.1 8B on our improved translations versus existing benchmark translations. As shown in Table~\ref{tab:avg_improvement}, we observe higher scores on our translations, providing additional evidence of enhanced translation quality. Below, we provide average difference between evaluating on our translations and existing ones (positive difference means results are higher on our translations); more detailed results are presented in Appendix~\ref{app:evals}.

\begin{table}[h]
\centering
\setlength{\tabcolsep}{4pt}
\begin{tabular}{lcr}
\toprule
\textbf{Benchmark} & \textbf{Languages} & \textbf{AVG $\Delta$} \\
\midrule
Winogrande & {\tiny UK, RO, SK, BG, LT, TR, EL, ET} & \textcolor{green!60!black}{+3.42\%} \\
ARC-Challenge & {\tiny UK, RO, SK, BG} & \textcolor{green!60!black}{+2.35\%} \\
Hellaswag & {\tiny UK, RO, SK, BG} & \textcolor{green!60!black}{+1.63\%} \\
MMLU & {\tiny UK, RO, SK, BG, LT, TR, EL} & \textcolor{green!60!black}{+0.94\%} \\
\bottomrule
\end{tabular}
\caption{Average improvement per benchmark across all languages}
\label{tab:avg_improvement}
\end{table}

We can observe gains across all studied benchmarks, with the most significant improvement in Winogrande, where preserving contextual connection between questions and answers together with removing the leakage of the correct answer through grammatical gender markers plays a crucial role in evaluation consistency. 

\begin{table}[h]
\centering
\setlength{\tabcolsep}{4pt}
\begin{tabular}{lr@{\hskip 12pt}lr}
\toprule
\textbf{Language} & \textbf{AVG $\Delta$} & \textbf{Language} & \textbf{AVG $\Delta$} \\
\midrule
Greek & \textcolor{green!60!black}{+3.89\%} & Romanian & \textcolor{green!60!black}{+2.09\%} \\
Ukrainian & \textcolor{green!60!black}{+2.7\%} & Slovak & \textcolor{green!60!black}{+1.66\%} \\
Turkish & \textcolor{green!60!black}{+2.65\%} & Estonian & \textcolor{green!60!black}{+1.63\%} \\
Lithuanian & \textcolor{green!60!black}{+2.6\%} & Bulgarian & \textcolor{green!60!black}{+1.37\%} \\
\bottomrule
\end{tabular}
\caption{Average improvement per language across all benchmarks}
\label{tab:avg_improvement_lang}
\end{table}

As shown, our proposed framework effectively enhances translation quality for multilingual benchmarks, leading to more reliable and accurate evaluations of language models across diverse languages.

\section{Conclusion}

We present a novel automated translation framework designed to enable rapid, high-quality translation of datasets and benchmarks while minimizing human intervention. Our integrated methods demonstrate substantial improvements on WMT24++ and FLORES benchmarks as measured by COMET scores, with T-RANK and USI achieving the strongest performance through test-time compute strategies. The quality of our translated benchmarks is further validated through COMET Quality Estimation scores and LLM-as-a-judge evaluations, which confirm meaningful improvements over existing translations. Notably, evaluation results of mid-sized models such as Gemma 3, Qwen 3 and Llama 3.1 on our improved translations yield higher scores compared to existing benchmark translations, providing additional evidence of enhanced translation quality. These findings demonstrate that our framework effectively balances translation accuracy, computational efficiency, and scalability, offering a practical solution for creating high-quality multilingual evaluation resources. Future work should explore adaptive method selection based on translation difficulty, integration of dedicated quality models, and comprehensive evaluation across open-weight models to further enhance the framework's capabilities on languages beyond Europe.

\section*{Limitations}

Our study has several limitations that warrant consideration. First, we employ LLM-based scoring rather than dedicated translation quality models like COMET for the Best-of-N selection process, which may affect the reliability of candidate ranking. Second, our approach applies uniform methods across all entries without automatically estimating translation difficulty per input, though adaptive method selection based on text complexity could potentially improve efficiency and quality. Machine translation benchmarks demonstrate that the most advanced and computationally expensive methods do not always yield superior results for shorter text sequences, particularly those not used in question-answering contexts. We defer to users of our framework to select methods based on their specific objectives and resource constraints. Third, we rely primarily on closed-source models for translation, with limited testing of open-weight alternatives. We hypothesize that our proposed methods would yield greater benefits for open-weight models, which typically demonstrate weaker performance in zero-shot translation settings, though this requires validation through comprehensive evaluation. Finally, while we focus on Eastern and Southern European languages, further research is needed to assess the generalizability of our framework across a broader range of low-resource languages with diverse linguistic characteristics.

\section*{Ethics Statement}

This work aims to improve multilingual benchmark quality to enable more equitable evaluation of language models across diverse languages. We acknowledge several ethical considerations. First, our reliance on closed-source models raises reproducibility concerns; we plan to extend evaluation to open-weight alternatives to promote accessibility. Second, while focusing on Eastern European languages addresses an important gap, many lower-resource languages remain underrepresented, and we encourage extending these methods to broader linguistic contexts. Third, we recognize that improved benchmarks may still be imperfect or subject to translation model bias; however, accurate multilingual evaluation is crucial for developing culturally aware and safer AI systems. Our automated translation approach avoids potential labor exploitation concerns associated with human translation. 

The code and translated benchmarks produced in this work will be released under permissible open licenses to enable community validation and reproducibility. Generative AI systems were employed exclusively for language assistance, including paraphrasing, spell-checking, and stylistic refinement of the authors' original content, as well as for generating boilerplate code. All core research contributions, experimental design, and findings represent the original work of the authors. Used benchmarks and datasets are publicly available under permissive licensing; any ethical considerations related to their use are discussed in the original publications.


\bibliography{custom}

\appendix

\section{Appendix}
\label{sec:appendix}

\subsection{Translation Examples}
\label{app:translation_examples}

In this subsection we will provide some translation examples showcasing common errors and how our proposed methods address them. Figure~\ref{fig:trank_correction} demonstrates how T-RANK identifies and corrects translation errors through competitive ranking. Unlike all other methods, we observe improved ability of the judge model to notice translation flaws when evaluating the translation in competitive manner. Table~\ref{fig:winogrande_examples} illustrates a common issue where translating questions and answers separately can leak the correct answer through grammatical gender markers, and shows how the correct approach preserves context during translation. We also identify systematic translation errors present in existing multilingual benchmark translations for MMLU and Hellaswag, complementing the Winogrande example shown in Table~\ref{fig:winogrande_examples}. These errors arise when questions and answer options are translated independently, without shared grammatical context.

Table~\ref{tab:mmlu_examples} documents four categories of errors found in MMLU translations produced by Global-MMLU and Okapi. Table~\ref{tab:hellaswag_examples} shows a concrete Hellaswag example where translating the question and answer options without the shared narrative context produces unnatural phrasing, grammatical errors, and broken cohesion between the question stem and the answer options.

\begin{figure*}[!ht]
\begin{prompt}[Translation Ranking (T-RANK) Correction Example]
\small
\textbf{Candidate 1}\\[3pt]
\textbf{Question:} \foreignlanguage{ukrainian}{Ви спостерігаєте вид, який народжує лише одне потомство за раз і має відносно тривалий життєвий цикл для свого розміру тіла. Що з наступного, ймовірно, також вірно для цього організму?}\\[3pt]
\textbf{Answers:}\\
1. \foreignlanguage{ukrainian}{Він живе в новозаселеному середовищі.}\\
2. \foreignlanguage{ukrainian}{Це водний організм.}\\
3. \foreignlanguage{ukrainian}{Він потребує відносно високої батьківської турботи про потомство.}\\
4. \foreignlanguage{ukrainian}{Вік, в якому потомство саме може народжувати, є відносно молодим.}\\[6pt]
\hrule\vspace{6pt}
\textbf{Candidate 2}\\[3pt]
\textbf{Question:} \foreignlanguage{ukrainian}{Ви спостерігаєте вид, який народжує лише одне потомство за раз і має відносно тривалий термін життя для свого розміру. Що з наведеного, ймовірно, також вірно для цього організму?}\\[3pt]
\textbf{Answers:}\\
1. \foreignlanguage{ukrainian}{Він живе в новозаселеному середовищі.}\\
2. \foreignlanguage{ukrainian}{Це водний організм.}\\
3. \foreignlanguage{ukrainian}{Він потребує відносно великої батьківської турботи про потомство.}\\
4. \foreignlanguage{ukrainian}{Вік, в якому потомство саме може народжувати, є відносно молодим.}\\[6pt]
\hrule\vspace{6pt}
\textbf{Candidate 3}\\[3pt]
\textbf{Question:} \foreignlanguage{ukrainian}{Ви спостерігаєте вид, який народжує лише одне потомство за раз і має відносно тривалий термін життя для свого розміру. Що з наведеного, ймовірно, також вірно для цього організму?}\\[3pt]
\textbf{Answers:}\\
1. \foreignlanguage{ukrainian}{Він живе в нещодавно колонізованому середовищі.}\\
2. \foreignlanguage{ukrainian}{Це водний організм.}\\
3. \foreignlanguage{ukrainian}{Він потребує відносно великої батьківської опіки за потомством.}\\
4. \foreignlanguage{ukrainian}{Вік, в якому потомство може самостійно народжувати, відносно молодий.}\\[6pt]
\hrule\vspace{6pt}
\textbf{Selected Translation}\\[3pt]
\textbf{Question:} \foreignlanguage{ukrainian}{Ви спостерігаєте вид, який народжує лише одне потомство за раз і має відносно тривалий термін життя для свого розміру. Що з наведеного, ймовірно, також вірно для цього організму?}\\[3pt]
\textbf{Answers:}\\
1. \foreignlanguage{ukrainian}{Він живе в нещодавно колонізованому середовищі.}\\
2. \foreignlanguage{ukrainian}{Це водний організм.}\\
3. \foreignlanguage{ukrainian}{Він потребує відносно великої батьківської опіки за потомством.}\\
4. \foreignlanguage{ukrainian}{Вік, в якому потомство може самостійно народжувати, відносно молодий.}\\[6pt]
\hrule\vspace{6pt}
\textbf{Reasoning}\\[3pt]
The final selected translation is generally accurate and conveys the original meaning well. However, there are minor improvements needed in terms of word choice and grammatical structure. The phrase \foreignlanguage{ukrainian}{`відносно тривалий термін життя'} could be simplified to \foreignlanguage{ukrainian}{`відносно довгий термін життя'} for better clarity. Additionally, the phrase \foreignlanguage{ukrainian}{`батьківської опіки за потомством'} can be improved to \foreignlanguage{ukrainian}{`батьківської опіки над потомством'} for grammatical correctness. Overall, the translation is good but requires these minor adjustments for optimal clarity and correctness.\\[6pt]
\hrule\vspace{6pt}
\textbf{Final Corrected Translation}\\[3pt]
\textbf{Question:} \foreignlanguage{ukrainian}{Ви спостерігаєте вид, який народжує лише одне потомство за раз і має відносно довгий термін життя для свого розміру. Що з наведеного, ймовірно, також вірно для цього організму?}\\[3pt]
\textbf{Answers:}\\
1. \foreignlanguage{ukrainian}{Він живе в нещодавно колонізованому середовищі.}\\
2. \foreignlanguage{ukrainian}{Це водний організм.}\\
3. \foreignlanguage{ukrainian}{Він потребує відносно великої батьківської опіки над потомством.}\\
4. \foreignlanguage{ukrainian}{Вік, в якому потомство може самостійно народжувати, відносно молодий.}
\end{prompt}
\caption{Translation Ranking (T-RANK) Correction Example}
\label{fig:trank_correction}
\end{figure*}

\begin{figure*}[!ht]
\begin{prompt}[Universal Self-Improvement (USI) Correction Example]
\small
\textbf{Candidate 1}\\[3pt]
\textbf{Question:} \foreignlanguage{ukrainian}{Ви спостерігаєте вид, який народжує лише одне потомство за раз і має відносно тривалий життєвий цикл для свого розміру тіла. Що з наступного, ймовірно, також вірно для цього організму?}\\[3pt]
\textbf{Answers:}\\
1. \foreignlanguage{ukrainian}{Він живе в новозаселеному середовищі.}\\
2. \foreignlanguage{ukrainian}{Це водний організм.}\\
3. \foreignlanguage{ukrainian}{Він потребує відносно високої батьківської турботи про потомство.}\\
4. \foreignlanguage{ukrainian}{Вік, в якому потомство саме може народжувати, є відносно молодим.}\\[6pt]
\hrule\vspace{6pt}
\textbf{Candidate 2}\\[3pt]
\textbf{Question:} \foreignlanguage{ukrainian}{Ви спостерігаєте вид, який народжує лише одне потомство за раз і має відносно тривалий термін життя для свого розміру. Що з наведеного, ймовірно, також вірно для цього організму?}\\[3pt]
\textbf{Answers:}\\
1. \foreignlanguage{ukrainian}{Він живе в новозаселеному середовищі.}\\
2. \foreignlanguage{ukrainian}{Це водний організм.}\\
3. \foreignlanguage{ukrainian}{Він потребує відносно великої батьківської турботи про потомство.}\\
4. \foreignlanguage{ukrainian}{Вік, в якому потомство саме може народжувати, є відносно молодим.}\\[6pt]
\hrule\vspace{6pt}
\textbf{Candidate 3}\\[3pt]
\textbf{Question:} \foreignlanguage{ukrainian}{Ви спостерігаєте вид, який народжує лише одне потомство за раз і має відносно тривалий термін життя для свого розміру. Що з наведеного, ймовірно, також вірно для цього організму?}\\[3pt]
\textbf{Answers:}\\
1. \foreignlanguage{ukrainian}{Він живе в нещодавно колонізованому середовищі.}\\
2. \foreignlanguage{ukrainian}{Це водний організм.}\\
3. \foreignlanguage{ukrainian}{Він потребує відносно великої батьківської опіки за потомством.}\\
4. \foreignlanguage{ukrainian}{Вік, в якому потомство може самостійно народжувати, відносно молодий.}\\[6pt]
\hrule\vspace{6pt}
\textbf{Selected Translation}\\[3pt]
\textbf{Question:} \foreignlanguage{ukrainian}{Ви спостерігаєте вид, який народжує лише одне потомство за раз і має відносно тривалий термін життя для свого розміру. Що з наведеного, ймовірно, також вірно для цього організму?}\\[3pt]
\textbf{Answers:}\\
1. \foreignlanguage{ukrainian}{Він живе в нещодавно колонізованому середовищі.}\\
2. \foreignlanguage{ukrainian}{Це водний організм.}\\
3. \foreignlanguage{ukrainian}{Він потребує відносно великої батьківської опіки за потомством.}\\
4. \foreignlanguage{ukrainian}{Вік, в якому потомство може самостійно народжувати, відносно молодий.}\\[6pt]
\hrule\vspace{6pt}
\textbf{Reasoning}\\[3pt]
<no reasoning used for USI method>\\[6pt]
\hrule\vspace{6pt}
\textbf{Final Corrected Translation}\\[3pt]
\textbf{Question:} \foreignlanguage{ukrainian}{Ви спостерігаєте вид, який народжує лише одне потомство за раз і має відносно \emph{тривалий цикл життя} для свого розміру. Що з наведеного, ймовірно, також вірно для цього організму?}\\[3pt]
\textbf{Answers:}\\
1. \foreignlanguage{ukrainian}{Він живе в нещодавно колонізованому середовищі.}\\
2. \foreignlanguage{ukrainian}{Це водний організм.}\\
3. \foreignlanguage{ukrainian}{Він потребує відносно великої батьківської опіки \emph{за} потомством.}\\
4. \foreignlanguage{ukrainian}{Вік, в якому потомство може самостійно народжувати, \emph{відносно молодий}.}
\end{prompt}
\caption{Universal Self-Improvement (USI) Correction Example}
\label{fig:usi_correction}
\end{figure*}

Table~\ref{tab:trank_positional_bias} reports the positional bias observed during the T-RANK ranking step on MMLU (EN$\to$UK). Even though the multi-round strategy rotates candidates across positions to mitigate bias, a residual preference for the candidate presented at input position~2 remains visible: it receives the lowest (best) average rank and holds rank~1 in all five most frequent rank combinations.

\begin{table*}[!ht]
\centering
\begin{tabular}{cccc}
\toprule
\textbf{Input position} & \textbf{Avg.\ rank} & \textbf{Top-5 rank combination (positions 1--5)} & \textbf{Occurrences} \\
\midrule
1 & 3.01 & (4,\ 1,\ 3,\ 2,\ 5) & 4731 \\
2 & \textbf{2.06} & (3,\ 1,\ 2,\ 4,\ 5) & 3141 \\
3 & 2.93 & (4,\ 1,\ 2,\ 3,\ 5) & 2997 \\
4 & 3.07 & (3,\ 1,\ 4,\ 2,\ 5) & 2600 \\
5 & 3.93 & (2,\ 1,\ 3,\ 4,\ 5) & 2428 \\
\midrule
\multicolumn{2}{l}{Equal ranks} & \multicolumn{2}{r}{6324} \\
\bottomrule
\end{tabular}
\caption{Positional bias in T-RANK ranking for MMLU (EN$\to$UK, $n{=}5$ candidates for $p{=}1$, translation model \emph{gpt-4o-mini}). Average rank per input position (lower = ranked better) and the five most frequent rank combinations. Position~2 receives the best average rank (2.06) and holds rank~1 in every top combination, confirming a residual bias towards the second candidate despite the multi-round rotation strategy.}
\label{tab:trank_positional_bias}
\end{table*}

\begin{table*}[!ht]
  \centering
  \begin{tabular}{p{0.45\textwidth}|p{0.45\textwidth}}
    \toprule
    \textbf{\xmark Bad Example (MuBench, Answer Leakage)} & \textbf{\cmark Good Example (Ours, Correct Translation)} \\
    \midrule
    \foreignlanguage{ukrainian}{Вони хвилювалися, що вино зіпсує ліжко та ковдру, але \_ не \textcolor{red}{було зіпсовано}.} &
    \foreignlanguage{ukrainian}{Вони хвилювалися, що вино зіпсує ліжко та ковдру, але \_ не \textcolor{green!60!black}{бу(-в/-ла/-ло/-ли)} зіпсовано.} \\[6pt]
    What does the blank \_ refer to? & What does the blank \_ refer to? \\[6pt]
    Option A: \foreignlanguage{ukrainian}{ковдра} & Option A: \foreignlanguage{ukrainian}{ковдра} \\
    Option B: \foreignlanguage{ukrainian}{ліжко} & Option B: \foreignlanguage{ukrainian}{ліжко} \\[6pt]
    Answer with A or B. & Answer with A or B. \\[6pt]
    Answer: \foreignlanguage{ukrainian}{ліжко} & Answer: \foreignlanguage{ukrainian}{ліжко} ? \\
    \bottomrule
  \end{tabular}
  \caption{Winogrande translation examples. Left (MuBench UK Winogrande): answer leakage where the correct answer ``\foreignlanguage{ukrainian}{ліжко}'' is implied grammatically in the question itself. Right (our translation): correct translation with verb morphology masking ``\foreignlanguage{ukrainian}{бу(-в/-ла/-ло/-ли)}'' to prevent answer leakage.}
  \label{fig:winogrande_examples}
\end{table*}

\begin{table*}[!ht]
\centering
\small
\setlength{\tabcolsep}{5pt}
\begin{tabular}{p{0.14\textwidth} p{0.40\textwidth} p{0.37\textwidth}}
\toprule
\textbf{Issue Type} & \textbf{Example} & \textbf{Impact} \\
\midrule
Semantic Drift &
  \textbf{Original:} ``relatively long lifespan'' \newline
  \textbf{Translated:} \foreignlanguage{ukrainian}{``життєвий цикл''} (life cycle) \newline
  \textbf{Should be:} \foreignlanguage{ukrainian}{``тривалість життя''} (lifespan) &
  Changes the question meaning; ``life cycle'' refers to developmental stages, not longevity \\[8pt]
Wrong Terminology &
  \textbf{Original:} ``aquatic organism'' \newline
  \textbf{Translated:} \foreignlanguage{ukrainian}{``водяний організм''} \newline
  \textbf{Should be:} \foreignlanguage{ukrainian}{``водний організм''} &
  \foreignlanguage{ukrainian}{``Водяний''} means ``watery'' (like soup), while \foreignlanguage{ukrainian}{``водний''} is the correct scientific term for aquatic \\[8pt]
Grammar Errors &
  \textbf{Original:} ``parental care for offspring'' \newline
  \textbf{Translated:} \foreignlanguage{ukrainian}{``турботи за потомством''} \newline
  \textbf{Should be:} \foreignlanguage{ukrainian}{``турботи про потомство''} &
  Incorrect preposition usage creates unnatural phrasing that may confuse native speakers \\[8pt]
Literal Translation &
  \textbf{Original:} ``Only I'' \newline
  \textbf{Translated:} \foreignlanguage{ukrainian}{``Тільки я''} \newline
  \textbf{Should be:} \foreignlanguage{ukrainian}{``Тільки I''} &
  The Latin numeral ``I'' is mis-read as the letter and rendered as the Ukrainian pronoun \foreignlanguage{ukrainian}{``я''} (I/me); the Roman numeral must be preserved in Latin script \\
\bottomrule
\end{tabular}
\caption{MMLU translation errors (Ukrainian) found in existing benchmarks (Global-MMLU, Okapi). Four recurring error categories with concrete examples and their impact on evaluation reliability.}
\label{tab:mmlu_examples}
\end{table*}

\begin{table*}[!ht]
  \centering
  \small
  \begin{tabular}{p{0.45\textwidth}|p{0.45\textwidth}}
    \toprule
    \textbf{\xmark~Bad Example (Okapi, Literal Translation)} & \textbf{\cmark~Good Example (Ours, Natural Phrasing)} \\
    \midrule
    \foreignlanguage{ukrainian}{Ми бачимо початковий екран заголовка. Ми бачимо }%
    \textcolor{red}{\foreignlanguage{ukrainian}{дівчинку біжать}}%
    \foreignlanguage{ukrainian}{ та робить високий стрибок, }%
    \textcolor{red}{\foreignlanguage{ukrainian}{пройшовши через перекладину}}%
    \foreignlanguage{ukrainian}{. Ми} &
    \foreignlanguage{ukrainian}{Ми бачимо титульний екран. Ми бачимо, як }%
    \textcolor{green!60!black}{\foreignlanguage{ukrainian}{дівчина біжить}}%
    \foreignlanguage{ukrainian}{, виконує стрибок у висоту і }%
    \textcolor{green!60!black}{\foreignlanguage{ukrainian}{перелазить через планку}}%
    \foreignlanguage{ukrainian}{. Ми} \\[5pt]
    \textit{\foreignlanguage{ukrainian}{Яке закінчення має найбільший сенс?}} &
    \textit{\foreignlanguage{ukrainian}{Яке закінчення має найбільший сенс?}} \\[4pt]
    A) \foreignlanguage{ukrainian}{бачимо близько 30 дівчат стрибають.} &
    A) \foreignlanguage{ukrainian}{бачимо, як близько 30 дівчат стрибають.} \\[2pt]
    B) \foreignlanguage{ukrainian}{бачимо, як вона закінчує перекладину та виконує рутину.} &
    B) \foreignlanguage{ukrainian}{бачимо, як вона закінчує стрибок над планкою і забирає свою рутину.} \\[2pt]
    C) \textcolor{red}{\foreignlanguage{ukrainian}{потім ми бачимо}}%
    \foreignlanguage{ukrainian}{ повтор і повільний повтор.} &
    C) \textcolor{green!60!black}{\foreignlanguage{ukrainian}{потім бачимо}}%
    \foreignlanguage{ukrainian}{ повтор і уповільнений повтор.} \\[2pt]
    D) \foreignlanguage{ukrainian}{бачимо, як дівчина кидає білий м'яч, і він летить прямо по полю.} &
    D) \foreignlanguage{ukrainian}{бачимо, як дівчина кидає білий м'яч, і він летить прямо через поле.} \\[5pt]
    \foreignlanguage{ukrainian}{Відповідь}: C &
    \foreignlanguage{ukrainian}{Відповідь}: C \\
    \midrule
    \textit{\textbf{Issue:} \textcolor{red}{\foreignlanguage{ukrainian}{``дівчинку біжать''}} mixes singular accusative with a plural verb; \textcolor{red}{\foreignlanguage{ukrainian}{``пройшовши через перекладину''}} is an awkward circumlocution for ``making it over the bar''; option~C inserts a disconnected subject \textcolor{red}{\foreignlanguage{ukrainian}{``ми''}} that breaks narrative flow.} &
    \textit{\textbf{Improvement:} \textcolor{green!60!black}{\foreignlanguage{ukrainian}{``дівчина біжить''}} restores correct agreement; \textcolor{green!60!black}{\foreignlanguage{ukrainian}{``перелазить через планку''}} is the natural sport phrase; \textcolor{green!60!black}{\foreignlanguage{ukrainian}{``потім бачимо''}} preserves the \foreignlanguage{ukrainian}{``Ми\ldots потім бачимо''} narrative flow.} \\
    \bottomrule
  \end{tabular}
  \caption{Hellaswag translation example (Ukrainian). Original English: ``We see an opening title screen. We see a girl run and perform a high jump and make it over the bar. We\ldots~--- which ending makes the most sense?''}
  \label{tab:hellaswag_examples}
\end{table*}

\subsection{Benchmark Translation Statistics}
\label{app:bench_stats}

In this subsection we provide statistics of selected benchmark translation, used splits and total number of samples used. We also provide information on models and methods used for final translation.

\begin{table*}[!t]
  \centering
  \begin{tabular}{lccccc}
    \hline
    \textbf{Benchmark} & \textbf{Train} & \textbf{Val} & \textbf{Dev} & \textbf{Test} & \textbf{All} \\
    \hline
    MMLU & -- & 1531 & 285 & 14042 & 15858 \\
    Winogrande & -- & 1267 & -- & -- & -- \\
    ARC-Challenge & 1119 & 299 & -- & 1172 & 2291 \\
    Hellaswag & -- & -- & -- & 10042 & 10042 \\
    \hline
    \textbf{Total} & 1119 & 3097 & 285 & 25556 & \textbf{29757} \\
    \hline
  \end{tabular}
  \caption{Number of examples in each benchmark split (-- indicates split not available or not used).}
  \label{tab:benchmark_stats}
\end{table*}

\begin{table*}[h]
\centering
\setlength{\tabcolsep}{5pt}
\begin{tabular}{lcccccccc}
\toprule
\textbf{Benchmark} & \textbf{UK} & \textbf{SK} & \textbf{RO} & \textbf{LT} & \textbf{ET} & \textbf{BG} & \textbf{EL} & \textbf{TR} \\
\midrule
MMLU & 4o/TR & 4o/TR & 4o/TR & 4o/USI & 4o/USI & 4o/TR & 4o/USI & 4o/USI \\
Hellaswag & 4o & 4o & 4o & G2F & G2F & G2F & G2F & G2F \\
ARC & 4o & G2F & G2F & G2F & G2F & G2F & G2F & G2F \\
Winogrande & 4o & G2F & G2F & G2F & G2F & G2F & G2F & G2F \\
\bottomrule
\end{tabular}
\caption{Model and method for benchmark translation. 4o = GPT-4o-mini, G2F = Gemini-2.0-Flash, TR = T-RANK. If method not specified, USI was used.}
\label{tab:best_models}
\end{table*}

\subsection{Detailed Evaluation Results}
\label{app:evals}

In this section we provide additional evaluation results on our own and existing translated benchmarks for analyzed languages. Tables~\ref{tab:eval_ukrainian} through~\ref{tab:eval_bulgarian_mubench} present performance comparisons across Ukrainian, Romanian, Slovak, Lithuanian, Bulgarian, Turkish, Greek and Estonian languages respectively, showing consistent improvements of our translations over existing benchmarks. Where available, we compare against Okapi/MuBench/Global-MMLU translations as baselines. For Bulgarian language we also compare against our own older translation done with the help of professional translators. With these results we confirm, that translation quality influences the evaluation results across various models with some exceptions to Winogrande. We note that professionally translated Winogrande to Bulgarian shows better results than our automated translation, which indicates that for some languages and benchmarks human intervention is still required to achieve the best quality, though we believe that with our methods the need for such intervention is significantly reduced.

\begin{table*}[!t]
  \centering  
\setlength{\tabcolsep}{4pt}
\begin{tabular}{l *{12}{>{\small}r}}
\toprule
& \multicolumn{3}{c}{\textbf{Hellaswag}} & \multicolumn{3}{c}{\textbf{Winogrande}} & \multicolumn{3}{c}{\textbf{ARC-Challenge}} & \multicolumn{3}{c}{\textbf{MMLU}} \\
\cmidrule(lr){2-4} \cmidrule(lr){5-7} \cmidrule(lr){8-10} \cmidrule(lr){11-13}
\textbf{Model} & O & Ot & $\Delta$\% & O & Ot & $\Delta$\% & O & Ot & $\Delta$\% & O & Ot & $\Delta$\% \\
\midrule
Gemma-3-12B-IT & 0.687 & 0.625 & \textcolor{green!60!black}{+6.2\%} & 0.580 & 0.619 & \textcolor{red}{-3.9\%} & 0.517 & 0.470 & \textcolor{green!60!black}{+4.8\%} & 0.614 & 0.603 & \textcolor{green!60!black}{+1.2\%} \\
Llama-3.1-8B & 0.570 & 0.540 & \textcolor{green!60!black}{+3\%} & 0.549 & 0.495 & \textcolor{green!60!black}{+5.5\%} & 0.431 & 0.416 & \textcolor{green!60!black}{+1.5\%} & 0.497 & 0.489 & \textcolor{green!60!black}{+0.9\%} \\
Gemma-3-4B-IT & 0.578 & 0.528 & \textcolor{green!60!black}{+5\%} & 0.540 & 0.501 & \textcolor{green!60!black}{+3.9\%} & 0.453 & 0.417 & \textcolor{green!60!black}{+3.7\%} & 0.453 & 0.444 & \textcolor{green!60!black}{+1\%} \\
Qwen3-8B-IT & 0.540 & 0.520 & \textcolor{green!60!black}{+2\%} & 0.569 & 0.546 & \textcolor{green!60!black}{+2.3\%} & 0.448 & 0.404 & \textcolor{green!60!black}{+4.4\%} & 0.619 & 0.599 & \textcolor{green!60!black}{+2\%} \\

\bottomrule
\end{tabular}
\caption{Ukrainian. O=Ours, Ot=Other (Okapi/MuBench/Global-MMLU)}
\label{tab:eval_ukrainian}
\end{table*}

\begin{table*}[!t]
  \centering 
\setlength{\tabcolsep}{4pt}
\begin{tabular}{l *{12}{>{\small}r}}
\toprule
& \multicolumn{3}{c}{\textbf{Hellaswag}} & \multicolumn{3}{c}{\textbf{Winogrande}} & \multicolumn{3}{c}{\textbf{ARC-Challenge}} & \multicolumn{3}{c}{\textbf{MMLU}} \\
\cmidrule(lr){2-4} \cmidrule(lr){5-7} \cmidrule(lr){8-10} \cmidrule(lr){11-13}
\textbf{Model} & O & Ot & $\Delta$\% & O & Ot & $\Delta$\% & O & Ot & $\Delta$\% & O & Ot & $\Delta$\% \\
\midrule
Gemma-3-12B-IT & 0.678 & 0.681 & \textcolor{red}{-0.2\%} & 0.661 & 0.589 & \textcolor{green!60!black}{+7.2\%} & 0.486 & 0.454 & \textcolor{green!60!black}{+3.2\%} & 0.628 & 0.614 & \textcolor{green!60!black}{+1.4\%} \\
Gemma-3-4B-IT & 0.567 & 0.567 & 0\% & 0.575 & 0.531 & \textcolor{green!60!black}{+4.5\%} & 0.412 & 0.394 & \textcolor{green!60!black}{+1.8\%} & 0.479 & 0.473 & \textcolor{green!60!black}{+0.6\%} \\
Llama-3.1-8B-IT & 0.576 & 0.581 & \textcolor{red}{-0.5\%} & 0.616 & 0.521 & \textcolor{green!60!black}{+9.5\%} & 0.358 & 0.357 & \textcolor{green!60!black}{+0.2\%} & 0.529 & 0.519 & \textcolor{green!60!black}{+0.9\%} \\
Qwen3-8B-IT & 0.538 & 0.537 & \textcolor{green!60!black}{+0.1\%} & 0.588 & 0.566 & \textcolor{green!60!black}{+2.3\%} & 0.368 & 0.359 & \textcolor{green!60!black}{+0.8\%} & 0.650 & 0.632 & \textcolor{green!60!black}{+1.8\%} \\
\bottomrule
\end{tabular}
\caption{Romanian. O=Ours, Ot=Other (Okapi/MuBench/Global-MMLU)}
\label{tab:eval_romanian}
\end{table*}

\begin{table*}[!t]
  \centering  
\setlength{\tabcolsep}{4pt}
\begin{tabular}{l *{6}{r}}
\toprule
& \multicolumn{3}{c}{\textbf{Winogrande}} & \multicolumn{3}{c}{\textbf{ARC-Challenge}} \\
\cmidrule(lr){2-4} \cmidrule(lr){5-7}
\textbf{Model} & Ours & Other & $\Delta$\% & Ours & Other & $\Delta$\% \\
\midrule
Gemma-3-12B-IT & 0.610 & 0.589 & \textcolor{green!60!black}{+2.1\%} & 0.539 & 0.509 & \textcolor{green!60!black}{+3\%} \\
Gemma-3-4B-IT & 0.554 & 0.517 & \textcolor{green!60!black}{+3.7\%} & 0.448 & 0.444 & \textcolor{green!60!black}{+0.4\%} \\
Llama-3.1-8B-IT & 0.547 & 0.515 & \textcolor{green!60!black}{+3.2\%} & 0.443 & 0.403 & \textcolor{green!60!black}{+4\%} \\
Qwen3-8B-IT & 0.555 & 0.559 & \textcolor{red}{-0.4\%} & 0.428 & 0.427 & \textcolor{green!60!black}{+0.1\%} \\
\bottomrule
\end{tabular}
\caption{Slovak. Other=MuBench/Okapi}
\label{tab:eval_slovak}
\end{table*}

\begin{table*}[!t]
  \centering 
\setlength{\tabcolsep}{4pt}
\begin{tabular}{l *{6}{r}}
\toprule
& \multicolumn{3}{c}{\textbf{Winogrande}} & \multicolumn{3}{c}{\textbf{MMLU}} \\
\cmidrule(lr){2-4} \cmidrule(lr){5-7}
\textbf{Model} & Ours & Other & $\Delta$\% & Ours & Other & $\Delta$\% \\
\midrule
Gemma-3-12B-IT & 0.631 & 0.557 & \textcolor{green!60!black}{+7.3\%} & 0.585 & 0.573 & \textcolor{green!60!black}{+1.3\%} \\
Gemma-3-4B-IT & 0.530 & 0.506 & \textcolor{green!60!black}{+2.3\%} & 0.433 & 0.410 & \textcolor{green!60!black}{+2.3\%} \\
Llama-3.1-8B-IT & 0.527 & 0.498 & \textcolor{green!60!black}{+2.9\%} & 0.431 & 0.417 & \textcolor{green!60!black}{+1.4\%} \\
Qwen3-8B-IT & 0.551 & 0.530 & \textcolor{green!60!black}{+2.1\%} & 0.553 & 0.541 & \textcolor{green!60!black}{+1.2\%} \\
\bottomrule
\end{tabular}
\caption{Lithuanian. Other=MuBench/Global-MMLU}
\label{tab:eval_lithuanian}
\end{table*}

\begin{table*}[!t]
  \centering  
\setlength{\tabcolsep}{4pt}
\begin{tabular}{l *{3}{r}}
\toprule
& \multicolumn{3}{c}{\textbf{Winogrande}} \\
\cmidrule(lr){2-4}
\textbf{Model} & Ours & Other & $\Delta$\% \\
\midrule
Gemma-3-12B-IT & 0.615 & 0.566 & \textcolor{green!60!black}{+4.9\%} \\
Gemma-3-4B-IT & 0.541 & 0.516 & \textcolor{green!60!black}{+2.5\%} \\
Llama-3.1-8B-IT & 0.543 & 0.509 & \textcolor{green!60!black}{+3.4\%} \\
Qwen3-8B-IT & 0.517 & 0.503 & \textcolor{green!60!black}{+1.4\%} \\
\bottomrule
\end{tabular}
\caption{Estonian. Other=MuBench}
\label{tab:eval_estonian}
\end{table*}

\begin{table*}[!t]
\centering
\setlength{\tabcolsep}{4pt}
\begin{tabular}{l *{6}{r}}
\toprule
& \multicolumn{3}{c}{\textbf{Winogrande}} & \multicolumn{3}{c}{\textbf{MMLU}} \\
\cmidrule(lr){2-4} \cmidrule(lr){5-7}
\textbf{Model} & Ours & Other & $\Delta$\% & Ours & Other & $\Delta$\% \\
\midrule
Gemma-3-12B-IT & 0.635 & 0.571 & \textcolor{green!60!black}{+6.5\%} & 0.587 & 0.581 & \textcolor{green!60!black}{+0.7\%} \\
Gemma-3-4B-IT & 0.583 & 0.514 & \textcolor{green!60!black}{+7\%} & 0.447 & 0.437 & \textcolor{green!60!black}{+1\%} \\
Llama-3.1-8B-IT & 0.573 & 0.518 & \textcolor{green!60!black}{+5.5\%} & 0.494 & 0.480 & \textcolor{green!60!black}{+1.4\%} \\
Qwen3-8B-IT & 0.545 & 0.571 & \textcolor{red}{-2.7\%} & 0.588 & 0.569 & \textcolor{green!60!black}{+1.8\%} \\
\bottomrule
\end{tabular}
\caption{Turkish. Other=MuBench/Global-MMLU}
\label{tab:eval_turkish}
\end{table*}

\begin{table*}[!t]
\centering
\setlength{\tabcolsep}{4pt}
\begin{tabular}{l *{6}{r}}
\toprule
& \multicolumn{3}{c}{\textbf{Winogrande}} & \multicolumn{3}{c}{\textbf{MMLU}} \\
\cmidrule(lr){2-4} \cmidrule(lr){5-7}
\textbf{Model} & Ours & Other & $\Delta$\% & Ours & Other & $\Delta$\% \\
\midrule
Gemma-3-12B-IT & 0.657 & 0.601 & \textcolor{green!60!black}{+5.6\%} & 0.593 & 0.580 & \textcolor{green!60!black}{+1.3\%} \\
Gemma-3-4B-IT & 0.598 & 0.518 & \textcolor{green!60!black}{+8.0\%} & 0.428 & 0.421 & \textcolor{green!60!black}{+0.7\%} \\
Llama-3.1-8B-IT & 0.593 & 0.520 & \textcolor{green!60!black}{+7.3\%} & 0.465 & 0.446 & \textcolor{green!60!black}{+2\%} \\
Qwen3-8B-IT & 0.581 & 0.528 & \textcolor{green!60!black}{+5.3\%} & 0.563 & 0.553 & \textcolor{green!60!black}{+1.1\%} \\
\bottomrule
\end{tabular}
\caption{Greek. Other=MuBench/Global-MMLU}
\label{tab:eval_greek}
\end{table*}

\begin{table*}[h]
\centering
\setlength{\tabcolsep}{3pt}
\begin{tabular}{l *{12}{>{\small}r}}
\toprule
& \multicolumn{3}{c}{\textbf{Hellaswag}} & \multicolumn{3}{c}{\textbf{Winogrande}} & \multicolumn{3}{c}{\textbf{ARC-Challenge}} & \multicolumn{3}{c}{\textbf{MMLU}} \\
\cmidrule(lr){2-4} \cmidrule(lr){5-7} \cmidrule(lr){8-10} \cmidrule(lr){11-13}
\textbf{Model} & {\centering Ours} & {\centering Other} & {\centering $\Delta$\%} & {\centering Ours} & {\centering Other} & {\centering $\Delta$\%} & {\centering Ours} & {\centering Other} & {\centering $\Delta$\%} & {\centering Ours} & {\centering Other} & {\centering $\Delta$\%} \\
\midrule
Gemma-3-12B-IT & 0.708 & 0.690 & \textcolor{green!60!black}{+1.8\%} & 0.643 & 0.677 & \textcolor{red}{-3.5\%} & 0.526 & 0.495 & \textcolor{green!60!black}{+3.1\%} & 0.605 & 0.619 & \textcolor{red}{-1.3\%} \\
Gemma-3-4B-IT & 0.590 & 0.569 & \textcolor{green!60!black}{+2.2\%} & 0.588 & 0.595 & \textcolor{red}{-0.7\%} & 0.430 & 0.410 & \textcolor{green!60!black}{+2\%} & 0.454 & 0.469 & \textcolor{red}{-1.5\%} \\
Llama-3.1-8B-IT & 0.539 & 0.528 & \textcolor{green!60!black}{+1.1\%} & 0.575 & 0.605 & \textcolor{red}{-3.1\%} & 0.394 & 0.367 & \textcolor{green!60!black}{+2.7\%} & 0.481 & 0.487 & \textcolor{red}{-0.6\%} \\
Qwen3-8B-IT & 0.536 & 0.518 & \textcolor{green!60!black}{+1.8\%} & 0.602 & 0.629 & \textcolor{red}{-2.7\%} & 0.414 & 0.394 & \textcolor{green!60!black}{+2\%} & 0.619 & 0.617 & \textcolor{green!60!black}{+0.2\%} \\
\bottomrule
\end{tabular}
\caption{Bulgarian. Other=INSAIT (professional translators)}
\label{tab:eval_bulgarian_insait}
\end{table*}

\begin{table*}[h]
\centering
\setlength{\tabcolsep}{3pt}
\begin{tabular}{l *{3}{r}}
\toprule
& \multicolumn{3}{c}{\textbf{Winogrande}} \\
\cmidrule(lr){2-4}
\textbf{Model} & Ours & Other & $\Delta$\% \\
\midrule
Gemma-3-12B-IT & 0.643 & 0.597 & \textcolor{green!60!black}{+4.6\%} \\
Gemma-3-4B-IT & 0.588 & 0.506 & \textcolor{green!60!black}{+8.2\%} \\
Llama-3.1-8B-IT & 0.575 & 0.505 & \textcolor{green!60!black}{+6.9\%} \\
Qwen3-8B-IT & 0.602 & 0.559 & \textcolor{green!60!black}{+4.3\%} \\
\bottomrule
\end{tabular}
\caption{Bulgarian. Other=MuBench}
\label{tab:eval_bulgarian_mubench}
\end{table*}

\subsection{Machine Translation Benchmark Evaluation Results}
\label{app:mt_benchmarks}

In this section, we test GPT-4o-mini and Gemini-2.0-Flash model on WMT and FLORES translation benchmarks, using both reference-based and reference-free COMET evaluation setups. As mentioned before, for reference-based evaluation we use Unbabel/XCOMET-XL model and for reference-free QE evaluation we use Unbabel/wmt23-cometkiwi-da-xl model. Tables~\ref{tab:wmt_qe_gpt4o} through~\ref{tab:flores_comet_gpt4o} present GPT-4o-mini results, while Tables~\ref{tab:wmt_qe_gemini} and~\ref{tab:wmt_comet_gemini} show Gemini-2.0-Flash performance.

\begin{table*}[!t]
\centering
\setlength{\tabcolsep}{3pt}
\begin{tabular}{l *{8}{>{\small}r}}
\toprule
\textbf{Method} & \textbf{EN$\to$UK} & \textbf{EN$\to$SK} & \textbf{EN$\to$RO} & \textbf{EN$\to$LT} & \textbf{EN$\to$ET} & \textbf{EN$\to$BG} & \textbf{EN$\to$TR} & \textbf{EN$\to$EL} \\
\midrule
Baseline & 0.726 & 0.741 & 0.882 & 0.741 & 0.788 & 0.751 & 0.718 & 0.802 \\
SC & 0.708 & 0.725 & 0.869 & 0.735 & 0.788 & 0.749 & 0.715 & 0.793 \\
BoN n=5 & 0.743 & 0.756 & 0.895 & 0.767 & 0.809 & 0.788 & 0.736 & 0.819 \\
USI n=5 & 0.750 & 0.759 & 0.899 & 0.766 & 0.806 & 0.791 & 0.733 & 0.817 \\
T-RANK n=5 & 0.739 & 0.745 & 0.885 & 0.749 & 0.799 & 0.776 & 0.729 & 0.811 \\
USI multi-prompt p=5 & \textbf{0.755} & \textbf{0.764} & \textbf{0.898} & \textbf{0.771} & \textbf{0.809} & \textbf{0.793} & \textbf{0.736} & \textbf{0.825} \\
T-RANK multi-prompt p=5 & 0.742 & 0.756 & 0.883 & 0.753 & 0.797 & 0.778 & 0.726 & 0.811 \\
\bottomrule
\end{tabular}
\caption{WMT QE (Quality Estimation) reference-free COMET scores for GPT-4o-mini}
\label{tab:wmt_qe_gpt4o}
\end{table*}

\begin{table*}[!t]
\centering
\setlength{\tabcolsep}{3pt}
\begin{tabular}{l *{8}{>{\small}r}}
\toprule
\textbf{Method} & \textbf{EN$\to$UK} & \textbf{EN$\to$SK} & \textbf{EN$\to$RO} & \textbf{EN$\to$LT} & \textbf{EN$\to$ET} & \textbf{EN$\to$BG} & \textbf{EN$\to$TR} & \textbf{EN$\to$EL} \\
\midrule
Baseline & 0.827 & 0.822 & 0.873 & 0.788 & 0.821 & 0.834 & 0.776 & 0.820 \\
SC & 0.821 & 0.817 & 0.869 & 0.790 & 0.822 & 0.834 & 0.834 & 0.821 \\
BoN n=5 & 0.844 & 0.837 & 0.884 & 0.805 & 0.838 & 0.852 & 0.791 & 0.836 \\
USI n=5 & 0.843 & 0.839 & 0.888 & 0.807 & 0.842 & 0.856 & 0.791 & 0.838 \\
T-RANK n=5 & 0.840 & 0.834 & 0.885 & 0.803 & 0.836 & 0.855 & 0.793 & 0.837 \\
USI multi-prompt p=5 & \textbf{0.849} & \textbf{0.847} & \textbf{0.891} & \textbf{0.817} & \textbf{0.848} & 0.862 & \textbf{0.803} & \textbf{0.844} \\
T-RANK multi-prompt p=5 & 0.845 & 0.841 & 0.887 & 0.812 & 0.843 & \textbf{0.862} & 0.798 & 0.841 \\
\bottomrule
\end{tabular}
\caption{WMT COMET reference-based scores for GPT-4o-mini}
\label{tab:wmt_comet_gpt4o}
\end{table*}

\begin{table*}[!t]
\centering
\setlength{\tabcolsep}{3pt}
\begin{tabular}{l *{8}{>{\small}r}}
\toprule
\textbf{Method} & \textbf{EN$\to$UK} & \textbf{EN$\to$SK} & \textbf{EN$\to$RO} & \textbf{EN$\to$LT} & \textbf{EN$\to$ET} & \textbf{EN$\to$BG} & \textbf{EN$\to$TR} & \textbf{EN$\to$EL} \\
\midrule
Baseline & 0.904 & 0.906 & 0.947 & 0.906 & 0.927 & 0.903 & 0.904 & 0.896 \\
SC & 0.902 & 0.908 & 0.950 & 0.911 & 0.935 & 0.902 & 0.903 & 0.896\\
BoN n=5 & 0.910 & 0.917 & 0.956 & 0.921 & 0.947 & 0.916 & 0.912 & 0.911\\
USI n=5 & 0.913 & 0.915 & 0.956 & 0.924 & 0.943 & 0.917 & 0.910 & 0.909\\
T-RANK n=5 & 0.899 & 0.906 & 0.952 & 0.909 & 0.932 & 0.911 & 0.903 & 0.902\\
USI multi-prompt p=5 & \textbf{0.919} & \textbf{0.921} & \textbf{0.961} & \textbf{0.932} & \textbf{0.948} & \textbf{0.922} & \textbf{0.918} & \textbf{0.912} \\
T-RANK multi-prompt p=5 & 0.905 & 0.912 & 0.954 & 0.912 & 0.935 & 0.912 & 0.908 & 0.904\\
\bottomrule
\end{tabular}
\caption{FLORES QE (Quality Estimation) reference-free COMET scores for GPT-4o-mini}
\label{tab:flores_qe_gpt4o}
\end{table*}

\begin{table*}[!t]
\centering
\setlength{\tabcolsep}{3pt}
\begin{tabular}{l *{8}{>{\small}r}}
\toprule
\textbf{Method} & \textbf{EN$\to$UK} & \textbf{EN$\to$SK} & \textbf{EN$\to$RO} & \textbf{EN$\to$LT} & \textbf{EN$\to$ET} & \textbf{EN$\to$BG} & \textbf{EN$\to$TR} & \textbf{EN$\to$EL}  \\
\midrule
Baseline & 0.937 & 0.938 & 0.939 & 0.906 & 0.894 & 0.943 & 0.920 & 0.920\\
SC & 0.937 & 0.937 & 0.936 & 0.911 & 0.902 & 0.940 & 0.916 & 0.923\\
BoN n=5 & 0.943 & 0.945 & 0.945 & 0.917 & 0.918 & 0.951 & 0.927 & 0.931\\
USI n=5 & 0.945 & 0.942 & 0.947 & 0.922 & 0.919 & 0.951 & 0.924 & 0.932\\
T-RANK n=5 & 0.938 & 0.937 & 0.943 & 0.910 & 0.903 & 0.947 & 0.920 & 0.929\\
USI multi-prompt p=5 & \textbf{0.947} & \textbf{0.946} & \textbf{0.949} & \textbf{0.928} & \textbf{0.918} & \textbf{0.952} & \textbf{0.931} & \textbf{0.934}\\
T-RANK multi-prompt p=5 & 0.940 & 0.943 & 0.944 & 0.915 & 0.907 & 0.950 & 0.925 & 0.932\\
\bottomrule
\end{tabular}
\caption{FLORES COMET reference-based scores for GPT-4o-mini}
\label{tab:flores_comet_gpt4o}
\end{table*}

We see, that on both datasets and evaluation types, USI method gives the best results. We also see, that using sampling single candidate from various prompts in both English and target language (2 English and 2 target languages prompts) gives additional boost in quality over single-prompt generation.

\begin{table*}[!t]
\centering
\setlength{\tabcolsep}{3pt}
\begin{tabular}{l *{8}{>{\small}r}}
\toprule
\textbf{Method} & \textbf{EN$\to$UK} & \textbf{EN$\to$SK} & \textbf{EN$\to$RO} & \textbf{EN$\to$LT} & \textbf{EN$\to$ET} & \textbf{EN$\to$BG} & \textbf{EN$\to$TR} & \textbf{EN$\to$EL} \\
\midrule
Baseline & 0.688 & 0.703 & 0.777 & 0.704 & 0.737 & 0.708 & 0.683 & 0.719 \\
SC & 0.687 & 0.706 & 0.777 & 0.706 & 0.736 & 0.709 & 0.685 & 0.722 \\
BoN n=5 & 0.687 & 0.707 & 0.780 & 0.710 & 0.743 & 0.713 & 0.689 & 0.723 \\
USI multi-prompt p=5 & \textbf{0.698} & \textbf{0.716} & \textbf{0.785} & \textbf{0.718} & \textbf{0.753} & \textbf{0.721} & \textbf{0.697} & \textbf{0.731} \\
T-RANK multi-prompt p=5 & 0.697 & 0.713 & 0.781 & 0.716 & 0.749 & 0.718 & 0.695 & 0.726 \\
\bottomrule
\end{tabular}
\caption{WMT QE (Quality Estimation) reference-free COMET scores for Gemini-2.0-flash}
\label{tab:wmt_qe_gemini}
\end{table*}

\begin{table*}[!t]
\centering
\setlength{\tabcolsep}{3pt}
\begin{tabular}{l *{8}{>{\small}r}}
\toprule
\textbf{Method} & \textbf{EN$\to$UK} & \textbf{EN$\to$SK} & \textbf{EN$\to$RO} & \textbf{EN$\to$LT} & \textbf{EN$\to$ET} & \textbf{EN$\to$BG} & \textbf{EN$\to$TR} & \textbf{EN$\to$EL} \\
\midrule
Baseline & 0.828 & 0.829 & 0.867 & 0.805 & 0.847 & 0.840 & 0.778 & 0.821 \\
SC & 0.826 & 0.834 & 0.868 & 0.808 & 0.845 & 0.841 & 0.779 & 0.824 \\
BoN n=5 & 0.828 & 0.836 & 0.870 & 0.814 & 0.852 & 0.843 & 0.785 & 0.823 \\
USI multi-prompt p=5 & \textbf{0.841} & 0.843 & 0.881 & 0.826 & \textbf{0.862} & 0.856 & 0.797 & 0.836 \\
T-RANK multi-prompt p=5 & 0.841 & \textbf{0.849} & \textbf{0.882} & \textbf{0.827} & 0.861 & \textbf{0.859} & \textbf{0.800} & \textbf{0.838} \\
\bottomrule
\end{tabular}
\caption{WMT COMET reference-based scores for Gemini-2.0-flash}
\label{tab:wmt_comet_gemini}
\end{table*}

In this case, for Gemini-2.0-Flash we observe that T-RANK multi-prompt method can outperform USI on most language pairs, although the difference is quite small. We believe that both methods have their strong and weak points -- while for datasets with shorter texts we recommends using USI, for benchmark translation we observed improved quality from using T-RANK method specifically, as shown in first subsection of appendix. Moreover, for Gemini-2.0-Flash model the ranking capabilities seem to be stronger, which may also contribute to better performance of T-RANK method. This supports our idea, that different methods may be more suitable for different models and use-cases.

\subsection{Translation Prompts}
\label{app:translation_prompts}

Table~\ref{tab:translation_prompts} provides the complete prompt templates used for all described translation methods. Figure~\ref{fig:judge_eval_prompt} provides the prompt for LLM-as-a-judge to compare our translations versus Global-MMLU as described in Table~\ref{tab:trank_mmlu_judge}.

\begin{table*}[h!]
\centering
\small
\begin{tabularx}{\textwidth}{lX}
\toprule
\textbf{Method} & \textbf{Prompt Template} \\
\midrule

Base Translation & 
\textbf{Instructions:} Imagine you're part of a team at an international education center that's revamping its exams for a global audience. Your job is to translate an English question and its answer options into <target\_language> so that students from <target\_language> schools can be evaluated too. Just provide the final translation---leave out any extra comments or explanations. Use language which is authentic for <target\_language> natives. Remember to keep the answer options connected to the question, using the same format as the original (a list for multiple choices or plain text for a single answer). Please do not translate valid code in any of the programming languages. 

\textbf{Original text:} \{"Original\_question": "<original\_question>", "Original\_answers": "<original\_answers>"\} 

\textbf{Output instructions:} Now, please give your final translation in <target\_language> exactly in this format, with only the translated content: \{"Question": "your\_translated\_question", "Answers": "translated\_answers"\} \\
\midrule

USI Judge & 
My task is to translate BENCHMARK questions with answers from English to <target\_language>. Your task is to evaluate if the response preserves the original question idea and to verify the correctness of declension and conjunction of words in the target language. 

The original text in English is: Question: <original\_question>, Answers: <original\_answers>

I have generated the following responses: <responses>

Combine the best features from responses to form the best response from grammatical and coherent points of view. Look for the next metrics:
* Quality of translation, including grammatical correctness
* Domain knowledge - were the terms correctly translated w.r.t to domain? Were coding terms or function names preserved (not translated)?
* Is the question text fully translated? Is the question idea preserved? 
* Are all the answer options/texts fully translated?
* Are the words written correctly? Are there any typos?

Output only the selected response: Question: selected question, Answer: selected answers \\
\midrule

T-RANK Ranking & 
My task is to translate BENCHMARK questions from English to <target\_language>. Your task is to rank my translations and select the best one. 

\textbf{Ranking criteria:} 

* Quality of translation
* Domain knowledge - were the terms correctly translated w.r.t to domain? Were coding terms or function names preserved (one would usually not translate coding operators or function names)?
* Is the question text fully translated? Is the question idea preserved? 
* Are all the answer options/texts fully translated?
* Are the words written correctly? Are there any typos?
* Are the declension and the conjunction of words written correctly in the target language if you connect answer options with the question?

\textbf{Original:} \{"original\_question": "<original\_question>", "original\_answers": <original\_answers>\}

\textbf{Candidates:} <responses>

\textbf{Instruction:} Select the best response (1st place = best). Correct if needed before output.

\textbf{Output:} Reasoning, then: \{"summary": "...", "final\_ranks": \{...\}, "rankings\_list": [...], "best\_translation": \{...\}\} \\
\midrule

Best-of-N Scoring & 
My task is to translate questions from English to <target\_language>. Score my translations 1-10 (10 = best).

\textbf{Original}: Question: <original\_question>, Answers: <original\_answers>

\textbf{Scoring metrics}: (1) Translation quality; (2) Question fully translated?; (3) All answers translated?; (4) Question idea preserved?; (5) Correct grammar?

\textbf{Responses}: <responses>

Output scores only: Response 1: score, Response 2: score, ..., Answers: [list of scores] \\

\bottomrule
\end{tabularx}
\caption{Translation prompt templates for different methods}
\label{tab:translation_prompts}
\end{table*}

\begin{figure*}[!ht]
\begin{prompt}[LLM-as-a-judge Quality Evaluation Prompt]
\textbf{Current User Query}

{"instruction": Your task is compare two translation from from English to Ukrainian and compare their advantages and disadvantages. Lastly, you have to pick one of three choices: A is better, B is better or they are equal.}

Original text (English)

<original\_text>

Translation A (Ukrainian)

<begin\_of\_translation\_A>
<output\_1>
<end\_of\_translation\_A> 

Translation B (Ukrainian)

<begin\_of\_translation\_B>
<output\_2>
<end\_of\_translation\_B> 

\textbf{Evaluation}

\textbf{Rules} 

You should compare the above two translations from English to Ukrainian based on your analysis of the user queries. You should first write down your analysis and the checklist that you used for the evaluation, and then provide your assessment according to the checklist. There are two choices to give your final assessment: ["A+", "B+",], which correspond to the following meanings:

- `A+`: Translation A is better than Translation B. 

- `B+`: Translation B is better than Translation A.

- `T=`: Translation A and Translation B are equally good. Tie.

\textbf{Output Format}

First, please output your analysis for each model translation from English to Ukrainian, and then summarize your assessment to three aspects: "reason A>B", and "reason B>A", and finally make your choice for the final assessment. Please provide your evaluation results in the following json format (starting with {{ and ending with }} ), with no quotes around the curly brackets by filling in the placeholders in []: 

analysis\_of\_A: <analysis of Translation A>, 

analysis\_of\_B: <analysis of Translation B>, 

reason\_of\_A\_equals\_B: <where Translation A and B perform equally well>,

reason\_of\_A\_better\_than\_B: <where Translation A is better than Translation B>, 

reason\_of\_B\_better\_than\_A: <where Translation B is better than Translation A>,  

choice: <A+, B+ or T=> 

Your translation should be a valid json dictionary following exactly this format.

\textbf{Your answer:}
\end{prompt}
\caption{LLM-as-a-judge Quality Evaluation Prompt}
\label{fig:judge_eval_prompt}
\end{figure*}

\end{document}